\documentclass[conference]{IEEEtran}
\IEEEoverridecommandlockouts
\usepackage{amsmath,amssymb,amsfonts}
\usepackage{algorithmic}
\usepackage{float}
\usepackage{algorithm}
\usepackage{graphicx}
\usepackage{textcomp}
\usepackage{xcolor}
\usepackage{multirow}
\usepackage{enumerate}
\usepackage{ctable}
\usepackage{subfig}
\usepackage{hyperref}

\usepackage[style=ieee]{biblatex}
\begin{document}

\title{Epsilon non-Greedy: A Bandit Approach for Unbiased Recommendation via Uniform Data}
\author{\IEEEauthorblockN{S.M.F. Sani}
\IEEEauthorblockA{\textit{Department of Computer Engineering} \\
\textit{Sharif University of Technology}\\
Tehran, Iran \\
s.feyzabadisani76@sharif.edu}
\and
\IEEEauthorblockN{Seyed Abbas Hosseini}
\IEEEauthorblockA{\textit{Department of Computer Engineering} \\
\textit{Sharif University of Technology}\\
Tehran, Iran \\
abbashossini@sharif.edu}
\and
\IEEEauthorblockN{Hamid R. Rabiee}
\IEEEauthorblockA{\textit{Department of Computer Engineering} \\
\textit{Sharif University of Technology}\\
Tehran, Iran \\
rabiee@sharif.edu}
}

\maketitle

\begin{abstract}
Often, recommendation systems employ continuous training, leading to a self-feedback loop bias in which the system becomes biased toward its previous recommendations. Recent studies have attempted to mitigate this bias by collecting small amounts of unbiased data. While these studies have successfully developed less biased models, they ignore the crucial fact that the recommendations generated by the model serve as the training data for subsequent training sessions. To address this issue, we propose a framework that learns an unbiased estimator using a small amount of uniformly collected data and focuses on generating improved training data for subsequent training iterations. To accomplish this, we view recommendation as a contextual multi-arm bandit problem and emphasize on exploring items that the model has a limited understanding of. We introduce a new offline sequential training schema that simulates real-world continuous training scenarios in recommendation systems, offering a more appropriate framework for studying self-feedback bias. We demonstrate the superiority of our model over state-of-the-art debiasing methods by conducting extensive experiments using the proposed training schema.
\end{abstract}


\section{Introduction}
Recommendation systems are built by training a model on a dataset $D$ that contains historical interactions between users and items. This dataset is typically obtained from a previously deployed recommendation algorithm; hence learning from this data may introduce bias into the model. The goal is to learn a model represented by parameters $\theta$ that can estimate the probability $p(r|a, c)$. Here, $r$ indicates whether a user will show interest in the suggested item $a$, and $c$ includes all relevant contextual information associated with the user-item interaction.
The optimal parameters $\theta$ are typically obtained through point estimation techniques. This is typically done by minimizing a loss function over the dataset:
\begin{equation}
\label{equation:typic_loss}
\theta^* = \underset{\theta}{argmin} \sum_{i=1}^{|D|} w_i l(r_i, \hat{r}_i(\theta))
\end{equation}
Here, $r_i$ represents the true user feedback, while $\hat{r}_i(\theta)$ represents the model's prediction. The term $w_i$ denotes the sample weight, which is often set as $\frac{1}{|D|}$, providing equal weight to each sample in the dataset.
However, relying solely on estimating parameter $\theta$ based on \eqref{equation:typic_loss} can result in a biased model towards the training data. 

This bias, known as the self-feedback loop \cite{chen2020bias}, persists even when an optimal algorithm is utilized for dataset collection. When the learned model encounters novel items or contexts not present in the training data, it often provides suboptimal recommendations with high confidence based on its acquired knowledge. As a result, these recommendations generate biased training data for subsequent iterations, thereby perpetuating bias in subsequent models.

Recent research focused on addressing the self-feedback loop in recommendation systems can be categorized into two groups. The first group, termed debiasing methods, encompasses methods that aim to acquire an unbiased estimation of parameter $\theta$ from the initially biased dataset. The second group, denoted as uncertainty-aware methods, emphasizes the inherent nature of continuous training in recommendation systems. These approaches strive to generate recommendations in a non-greedy way that produce less biased training data for subsequent training rounds.

Debiasing methods aim to mitigate bias in recommendation models trained on biased datasets. One common approach to achieve unbiased models is by incorporating Inverse Propensity Scores (IPS) as sample weights in \eqref{equation:typic_loss} \cite{sun2019debiasing, saito2020asymmetric, saito2020unbiased}. However, these methods may face challenges associated with high variance, especially when the data collection algorithm differs significantly from the learning algorithm. Recent studies propose methods that leverage information from uniformly recommended items to learn unbiased models \cite{liu2020general,chen2021autodebias,liu2023bounding}. However, the amount of uniformly collected interactions in recommendation systems is often limited due to its negative impact on user satisfaction. The well-known epsilon-greedy algorithm exemplifies such an approach by utilizing a small amount of randomly suggested items to learn an unbiased model. Nonetheless, debiasing methods often prioritize recommending items with the highest expected feedback in a greedy manner without considering the impact on the composition of subsequent training datasets.

Uncertainty-aware methods are designed to optimize long-term user satisfaction by generating recommendations that contribute to the creation of enhanced training data. These methods typically integrate exploration mechanisms that prioritize items with higher uncertainty, ensuring that the model is exposed to recommendations it has limited knowledge about. 
This helps mitigate bias during continuous training iterations \cite{guo2020deep, jeunen2020joint, du2021exploration}. However, these methods overlook the unbiased nature of uniformly collected data as a valuable source for debiasing in recommendation systems.

To address these limitations, we introduced the Epsilon non-Greedy (EnG) framework, which combines the advantages of both groups above. Our EnG framework overcomes the limitations by achieving unbiased recommendations through learning on biased datasets and generating recommendations that can serve as less biased training data for subsequent training iterations, ultimately maximizing long-term user satisfaction.

Given the widespread use of deep neural networks in state-of-the-art recommendation algorithms \cite{cheng2016wide,he2017neural,xue2019deep}, we adopt them as our framework's underlying backbone. To achieve an unbiased recommendation system, we propose a teacher-student architecture with a novel training loss function that effectively leverages information from a small quantity of uniformly collected data. To incorporate systematic exploration into our framework, we view the recommendation as a contextual multi-arm bandit problem. We integrate Thompson sampling into our framework by employing the dropout technique, enabling us to recommend items the model has limited knowledge about. By incorporating this approach, the resulting interactions serve as improved training data for subsequent training sessions, ultimately enhancing long-term user satisfaction.

The conventional training schema commonly employed in recent studies does not accurately reflect the debiasing capabilities of recommendation systems. We propose a novel offline sequential training schema that simulates the continuous training process observed in real-world recommendation systems. Using this training schema, we perform comprehensive experiments and demonstrate that our proposed framework exhibits superior debiasing power compared to state-of-the-art methods. 
This paper presents several key contributions:
\begin{enumerate}[I]
    \item 
    \textbf{EnG Framework:} We introduce the EnG framework, which enables unbiased recommendations and generates training data with a reduced bias for subsequent training iterations. By utilizing a teacher-student architecture trained with the proposed loss function, the framework facilitates the development of an unbiased learner. Additionally, the incorporation of Thompson sampling through the dropout technique allows the model to explore items for which it lacks certainty, effectively breaking the self-feedback loop and enhancing long-term recommendation performance.
    \item
    \textbf{Sequential Training Schema}: We introduce a sequential training schema that closely aligns with real-world recommendation scenarios. This schema allows for evaluating debiasing capabilities in recommendation algorithms by simulating the continuous training process observed in real-world systems.
    \item
    \textbf{Experimental Evaluation}: Extensive experimentation is conducted on two popular real-world datasets. The results demonstrate the superior debiasing power of the proposed EnG framework, surpassing state-of-the-art methods.
\end{enumerate}

\section{Related Work}
This section summarizes prior research endeavors to tackle the problem of self-feedback loop bias in recommendation systems. The existing literature can be broadly classified into two main categories: methods that primarily aim to achieve unbiased estimators and techniques that incorporate uncertainty to enhance long-term user satisfaction by improving the quality of training data for subsequent training iterations.

\subsection{Debiasing Methods}
Current research endeavors aiming to obtain unbiased estimators from biased data can be categorized into two main groups. The first group primarily uses inverse propensity scores, while the second concentrates on leveraging a limited quantity of collected unbiased data to learn an unbiased model.

\subsubsection{Inverse Propensity Score}
Propensity scores are crucial in addressing self-feedback loop bias in recommendation systems. These scores represent the probability of observing a particular data point in the dataset and are typically determined by the recommendation algorithm. Methods that leverage propensity scores assign weights to each sample based on their inverse propensity, enabling the estimation of unbiased loss function of interest even when calculated on biased data \cite{schnabel2016recommendations}. 
The utilization of propensity scores has demonstrated improvements in the performance of matrix factorization methods \cite{liang2016causal,schnabel2016recommendations}. A recent study used propensity scores to predict users' preferences from Missing-Not-at-Random (MNAR) implicit feedback \cite{saito2020unbiased}. 

Furthermore, in seeking unbiased estimation for loss function of interest, researchers have also explored using propensity scores in conjunction with positive-unlabeled learning techniques \cite{saito2020asymmetric}.

Inverse-propensity-based estimators, commonly used for unbiased point estimation, often face the issue of high variance when there are substantial disparities between the recommendation algorithm and the data collection algorithm. To mitigate this challenge, several solutions, including self-normalization, clipped, and doubly robust estimators, have been proposed \cite{swaminathan2015self,saito2020unbiased,wang2019doubly}.

\subsubsection{Uniformly Collected Data}
Uniformly collected data refers to a dataset where recommendations are made with equal probabilities for different items. This data type provides an opportunity to estimate user preferences without the influence of self-feedback loop bias, resulting in an unbiased estimator \cite{yuan2019improving}. However, collecting a large amount of uniform data is impractical due to potential negative impacts on user satisfaction and business revenue.
The epsilon greedy algorithm is a fundamental approach that leverages small amounts of uniformly collected data. This algorithm incorporates a random item selection strategy with a probability of epsilon, while the remaining items are selected greedily based on their expected rewards. A recent study proposed a multi-task objective that jointly factors the model trained on biased data with the model trained on uniformly collected data \cite{bonner2018causal}. Another recent study presents an alternative approach employing knowledge distillation methods for counterfactual recommendations, specifically classifying them into four categories: label-based, sample-based, feature-based, and structure-based distillation \cite{liu2020general}. Our proposed approach is closely aligned with label-based distillation, particularly regarding the loss function utilized.
Influence functions were also used to assign weights directly to training samples \cite{yu2020influence}. 


A recent study proposed a general framework that addresses different biases using a meta-learning algorithm to obtain sample weights and provide unbiased estimations of the loss function of interest \cite{chen2021autodebias}.
Theoretical guarantees also support the effectiveness of these debiasing strategies; causal diagrams have been employed to model biased and unbiased feedback generation processes in recommendation systems \cite{liu2021mitigating}. A debiasing strategy based on information bottleneck has been proposed by identifying the confounding bias as the disparity between the two diagrams. A recent research investigation has provided upper bounds for the unbiased loss function of interest, encompassing both a generalization error and a separability-based bound. Building upon these bounds, a novel debiasing approach named debiasing approximate upper bound (DUB) has been introduced \cite{liu2023bounding}.
Recent studies also have demonstrated the potential benefits of integrating inverse propensity scores alongside uniformly collected data \cite{yuan2019improving,wang2021combating}. 

\subsection{Uncertainty Aware Methods}
Uncertainty-aware methods aim to address the self-feedback loop bias by approximating the posterior distribution of model parameters $p(\theta|D)$ or estimating its variance since it is a way to quantify the uncertainty associated with the model's predictions. Considering the uncertainty in the model's predictions, these methods promote exploratory recommendations to capture user interests and break the self-feedback loop. As the model explores diverse recommendations, the uncertainty decreases (convergence of the posterior distribution) and tends to exploit its learned knowledge more effectively.
The contextual multi-arm bandit problem has emerged as a widely adopted framework for an uncertainty-aware recommendation. A notable example is the LinUCB algorithm, which leverages contextual information to make informed recommendations \cite{li2010contextual}. In a related study, a factorization-based bandit algorithm that incorporates low-rank matrix completion by incrementally constructing a matrix representing user-item preferences has been proposed \cite{wang2017factorization}. 
Thompson sampling helps balance the exploration/exploitation trade-off in recommendation systems \cite{thompson1933likelihood}. 
This approach allows for a more principled exploration of the recommendation. Yarin Gal and Zoubin Ghahramani contributed significantly by providing a probabilistic interpretation of dropout in deep learning models. They developed a theoretical framework casting dropout training in deep neural networks as approximate Bayesian inference in deep Gaussian processes \cite{gal2016dropout}. 
 A recent study viewed the recommendation system as a contextual multi-arm bandit problem.  They compared different techniques, including bootstrapping, dropout, and a hybrid method, for drawing samples from the posterior distribution in their model \cite{guo2020deep}. Furthermore, the authors investigated various exploration mechanisms. They compared the performance of the epsilon-greedy, Thompson sampling, and upper confidence bound methods in terms of their effectiveness as exploration strategies. A study proposed a comprehensive review of Thompson sampling to balance the exploration/exploitation trade-off \cite{riquelme2018deep}. In addition, some studies investigate the usage of propensity scores alongside exploration mechanisms. For instance, learnable propensity weights are employed to achieve unbiased estimations in the REINFORCE algorithm \cite{chen2019top,williams1992simple}.  

\section{Epsilon non-Greedy}
In this section, we introduce our framework, referred to as Epsilon Non-Greedy (EnG), which is specifically developed to accomplish two key objectives: (1) establish an unbiased recommendation system by effectively utilizing a limited quantity of uniformly collected data and (2) generate recommendations that contribute to the creation of high-quality training data for subsequent training iterations. We make certain assumptions regarding the recommendation system. We consider a scenario where the system recommends only one item at a time to the user. Additionally, we model the user's feedback as a binary random variable, denoted as $r$, where $r=1$ indicates that the user likes the recommended item, while $r=0$ shows otherwise. Furthermore, we assume that the contextual information vector, represented as $c$, is sufficiently comprehensive to encompass all relevant information required for the recommendation problem. Therefore, estimating the probability $p(r=1|a, c)$ to address the recommendation task effectively is sufficient. 

We propose a teacher-student architecture accompanied by a novel loss function to effectively harness the information in the uniformly collected data. This component of our framework, which shares similarities with using random interactions in the epsilon-greedy algorithm to learn an unbiased model, is called the "epsilon" component. Recognizing the significance of recommended items in the continuous training process, we incorporate Thompson sampling into the EnG framework. By utilizing Thompson sampling, the framework recommends items based on their probability of being the optimal choice. This deviates from the greedy approach of selecting items with the highest expected instant reward. This probabilistic recommendation strategy is called our framework's "non-greedy" component.

To train the EnG framework, we utilize the historical interactions logged in a dataset denoted as $D = \{(r_i, a_i, c_i)\}_{i}$. This dataset can be divided into two main parts. The first part, denoted as $D^r$, consists of a relatively small amount of data collected using a uniform policy. This data is obtained by suggesting items to users with equal probability. The second part, denoted as $D^b$, comprises the logs of interactions between users and items that were selected by the recommendation algorithm itself. In the recommendation problem, the user-item interactions are often sparse, meaning that each user typically interacts with only a few items. Consequently, numerous user-item pairs remain unobserved in the dataset, denoted as $D^u$. In the subsequent sections, we will provide a detailed explanation of each framework component.

\subsection{Epsilon}
We propose a teacher-student architecture to unleash the unbiased information of $D^r$. The teacher model is specifically trained on the uniformly suggested items in $D^r$ to learn unbiased recommendations. Its role is to distill and transfer this knowledge to the student model, enabling the student to make less biased predictions. 
Fig \ref{fig:proposed-method} illustrates the proposed architecture of the EnG framework, which employs feedforward neural networks as the backbone for both the teacher and student models. The training process starts by training the teacher network solely on $D^r$. The loss function used for training the teacher network is defined as follows:
\begin{figure*}[t]
\centering
  \includegraphics[width=.75\linewidth]{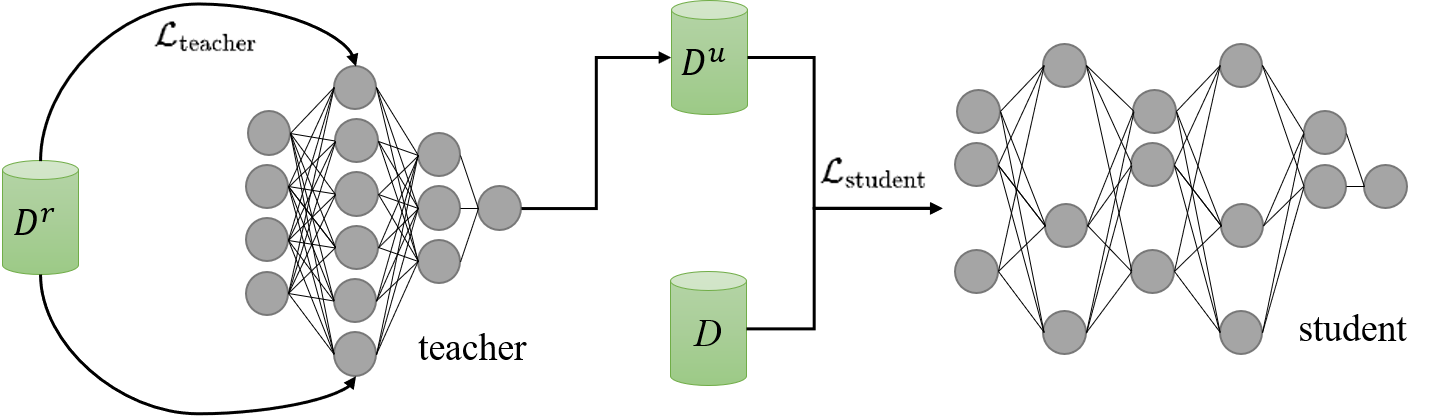}
\caption{(Left) The smaller teacher network is trained via  \eqref{equation:teacher_loss}. (Right) The student network is trained via \eqref{equation:student_loss}.}
\label{fig:proposed-method}
\end{figure*}
\begin{equation}
\label{equation:teacher_loss}
    \mathcal{L}_{t} = \frac{1}{|D^r|} \sum_{i=1}^{|D^r|} l(\hat{r}_i^{t}, r_i) + \lambda_{t} \mathcal{R}(\theta_{t})
\end{equation}
Where $\hat{r}_i^{t}$ represents the predicted reward by the teacher network for the $i$-th sample. The true reward is denoted as $r_i$. The loss function $l$ is employed to quantify the discrepancy between predicted and true rewards. We choose Binary Cross Entropy (BCE) as it is suitable for binary classification tasks. $\mathcal{R}$ represents the regularization term. The parameters of the teacher network are represented by $\theta_{t}$, and the hyperparameter $\lambda_{t}$ controls the regularization strength.
After the convergence of the teacher network, the student network is trained with the guidance of the teacher network using the following loss function:
\begin{equation}
\label{equation:student_loss}
    \begin{split}
        &\mathcal{L}_{s} = \frac{1}{|D|} \sum_{i=1}^{|D|} l(\hat{r}_i^{s}, r_i) +
        \frac{\gamma_{\text{reg}}}{|D^u|} \sum_{i=1}^{|D^u|} l_{\text{reg}}(\hat{r}_i^{s}, \hat{r}_i^{t}) + \lambda_{s} \mathcal{R}(\theta_{s})\\
    \end{split}
\end{equation}
The first term involves training the student network on all available data $D$. However, relying solely on this term could introduce bias into the model due to the majority of data being non-uniformly collected. To mitigate this bias, we introduce the second term, which leverages the knowledge of the teacher network to establish an unbiased estimator. The student network is regularized by encouraging it to have similar predictions to the teacher network on unobserved data $D^u$. Unobserved data is chosen as it provides a large amount of data for extensive training and shares similar unbiased characteristics with the uniformly collected data. The third term aims to regularize the weights of the student network to prevent overfitting. The $\gamma_{\text{reg}}$ is used to control regularization strength in \eqref{equation:student_loss}. 
To ensure that the student's predictions are close to the teacher's predictions, we consider two types of loss functions ($l_{\text{reg}}$). The first group of loss functions treats the predictions as logits and measures the discrepancy between these predictions. Examples of such loss functions include squared error and absolute error. The second group of loss functions considers the predictions as parameters of Bernoulli distributions, which effectively model the probability of obtaining a reward based on a given action and context. Notable examples of such loss functions include the Kullback-Leibler (KL) divergence and Jeffreys divergence, which represents a symmetric variant of the KL divergence.

Our proposed solution addresses the challenges of training a high-capacity model with limited uniform data by training the teacher and student models separately. The teacher model is initially trained with constrained capacity, $\theta_{t} \in \mathbb{R}^p, \theta_{s}\in \mathbb{R}^q$, where $p << q$, to mitigate variance and stabilize its recommendations. This ensures consistent and reliable knowledge transfer from the teacher to the student, even with limited uniformly collected data. By adopting this approach, the student model can benefit from the teacher's guidance while maintaining its capacity for accurate recommendations. This framework effectively enables the student model to learn from available data while leveraging the insights and expertise of the teacher model.
\subsection{Non-Greedy}
To this point, utilizing the proposed epsilon component of the framework has enabled us to recommend items with the highest expected instant reward.
\begin{equation}
\underset{a \in \mathcal{A}}{\arg \max} \; \mathbb{E}[r=1|a, c]
\end{equation}
However, this strategy is considered greedy as it continually selects items that maximize the expected instant reward. To evaluate the impact of recommendations on continuous training data, we can rewrite the probability $p(r=1|a, c)$ using Bayes' rule:
\begin{equation}
p(r=1|a, c) = \int p(r=1|a, c, \theta) p(\theta|D) d\theta
\end{equation}
By accessing this posterior distribution, a non-greedy item selection strategy emerges. Instead of selecting items solely based on expected instant reward, we can recommend items in proportion to the probability of them being optimal. This can be expressed as:
\begin{equation}
p(a=a^*|c) = \int \mathbb{I}[\mathbb{E}[r|a, c, \theta] = \underset{a'}{\max} \mathbb{E}[r|a', c, \theta]] p(\theta|D)d\theta
\end{equation}
In practice, obtaining the exact posterior distribution $p(\theta|D)$ and calculating the integral is intractable in many real-world scenarios. However, Thompson sampling provides an alternative technique. Instead of directly evaluating the integral, Thompson sampling involves sampling model parameters from the posterior distribution and selecting the best item based on these drawn parameters.
Thompson sampling allows for systematic exploration of the recommendation problem. Thompson sampling breaks the self-feedback loop by choosing items non-greedily using the drawn parameters. Initially, when the training begins, the posterior distribution is relatively flat, indicating uncertainty in the model parameters. Samples drawn from this distribution explore a wide range of parameter values, leading to the exploration of various items. As the training progresses, the posterior distribution becomes more peaked, indicating increased confidence in the learned parameters. Consequently, samples drawn from the posterior distribution tend to cluster around the most probable parameter values, prioritizing the exploitation of the learned knowledge.

To draw samples from the posterior distribution of neural network parameters, the study conducted by Yarin Gal and Zoubin Ghahramani offers valuable insights. They demonstrate that using the dropout technique during the inference stage of a neural network can be interpreted as obtaining model predictions based on samples drawn from the posterior distribution of the model parameters \cite{gal2016dropout}. Dropout is a computationally efficient approach, making it practical to implement within recommendation systems without significant computational overhead.
Therefore, the student network employs the dropout technique during the inference process  (Fig \ref{fig:proposed-method}). This allows for a balanced approach between exploration and exploitation in the recommendations made by the student network. By incorporating a teacher-student architecture and utilizing Thompson sampling during the recommendation process, the proposed framework effectively addresses the recommendation algorithm's short-term and long-term objectives.

\subsection{Sequential Training Algorithm}
\label{section:sequential_trainnig_schema}
Recent studies and endeavors aiming to disrupt the self-feedback loop in recommendation systems through uniformly collected data have predominantly adopted a conventional training approach. This approach involves partitioning the available uniform data into three distinct subsets: a small portion for model training, another portion for validation and hyperparameter tuning, and a final portion for unbiased evaluation.
However, the conventional training approach fails to fully capture the continuous training nature of real-world recommendation systems. Evaluating the model solely based on this training schema may not accurately reflect its long-term performance. We propose a sequential training schema specifically tailored for training recommendation systems to overcome this limitation and better simulate real-world scenarios.

The proposed training algorithm adopts a sequential process wherein the training data, comprising uniformly collected and biased logged data, is divided into $M$ batches. In each training round, the model predicts scores for all items in the batch. A proportion $\rho$ of items with the highest predicted scores are selected as chosen recommendations and included in the model's training data. This sequential training schema aims to emulate the behavior of recommendation systems in real-world settings, where collected interactions often consist of recommendations suggested by the model itself. By incorporating this sequential training schema, the algorithm aims to effectively integrate the continuous training aspect into the model learning process.

The complete training algorithm, presented in Algorithm \ref{algorithm:epsilon-non-greedy}, outlines the step-by-step procedure for implementing the proposed Epsilon non-Greedy framework. In the 7th line of the algorithm, we incorporate the dropout technique to obtain predictions. This utilization of dropout introduces additional diversity in the selected training data, thereby enhancing the exploration aspect of the dataset generation process for our recommendation model. When a method solely prioritizes learning an unbiased algorithm without considering the composition of subsequent training data, it tends to predominantly recommend items for which it already possesses knowledge. Consequently, the quality and diversity of the model's training data diminish, leading to a bias toward recommendations generated by the method during subsequent training iterations.

\begin{algorithm}[t]
\caption{Epsilon Non-Greedy}
\label{algorithm:epsilon-non-greedy}
\begin{algorithmic}[1]
\REQUIRE $D^r$, $D^b$, $D^u$, $M$, $\rho$, $\lambda_{\text{teacher}}$, $\lambda_{\text{student}}$, $\gamma_{\text{student}}$

\STATE Split $D^r$ into $D^r_{\text{train}}$, $D^r_{\text{validation}}$, $D^r_{\text{test}}$
\STATE Divide $D^r_{\text{train}}$, $D^b$ into $M$ equal batches ($d^r_i$, $d^b_i$)
\STATE Initialize $\theta_{\text{student}}^0$
\STATE $\mathcal{S}^b \leftarrow \{\}$
\STATE $\mathcal{S}^r \leftarrow \{\}$

\FOR {$i=1$ to $M$}
    \STATE $\text{scores} \leftarrow f_{\theta_{\text{student}}^{i-1}}(d^b_i)$
    \STATE Sort $d^b_i$ based on scores in descending order
    \STATE $\text{winners} \leftarrow$ top $\rho$ portion of $d^b_i$
    \STATE $\mathcal{S}^b \leftarrow \mathcal{S}^b \cup \text{winners}$
    \STATE $\mathcal{S}^r \leftarrow \mathcal{S}^r \cup d^r_i$
    \STATE Initialize $\theta_{\text{teacher}}^i$
    \STATE Initialize $\theta_{\text{student}}^i$
    \STATE  $\theta_{\text{teacher}}^i \leftarrow$ update on $\mathcal{S}^r$ using \eqref{equation:teacher_loss}
    \STATE $\theta_{\text{student}}^i \leftarrow$ update on $\mathcal{S}^r, \mathcal{S}^b, D^u$ using \eqref{equation:student_loss}
\ENDFOR

\RETURN $\theta_{\text{student}}^M$
\end{algorithmic}
\end{algorithm}

\section{Empirical Evaluation}
This section comprehensively evaluates the performance of the EnG framework compared to state-of-the-art methods. Firstly, we demonstrate the effectiveness of our teacher-student architecture in achieving unbiased recommendations using limited uniformly collected data through conventional and sequential training. Secondly, we investigate the potential enhancement of the proposed method by integrating Thompson sampling into the recommendation process, particularly in reducing biases over extended durations. Additionally, an ablation study is conducted to analyze the impact of important hyperparameters on the performance of EnG. The code can be accessed via \href{https://github.com/FeyzabadiSani/Epsilon-nonGreedy/}{github.com/FeyzabadiSani/Epsilon-nonGreedy/}

\subsubsection{Datasets}
This section provides an overview of the datasets used in our experiments. The statistical characteristics of these datasets, including the positive sample ratio (PR), are presented in Table \ref{table:dataset_statistics}.

 \textbf{YahooR3} \cite{marlin2009collaborative}: Comprises user ratings of songs on a scale ranging from 1 to 5. The dataset consists of interactions between 15,400 users and 1,000 songs. It is divided into two parts: the first part consists of user ratings collected during regular interactions with Yahoo music services, which can be considered as the biased portion of the dataset. The second part consists of user ratings on randomly selected songs, which can be considered as the uniform portion of the dataset.
 
 \textbf{Coat} \cite{schnabel2016recommendations}: It includes ratings of 290 users on 300 different coats, using a scale ranging from 1 to 5. Each user is initially asked to rate 24 coats based on their personal interests. Additionally, a further 16 coats are randomly selected for the user to rate, constituting the dataset's uniform data portion.

The datasets are binarized based on user ratings, representing the sparsity of user feedback. Ratings of 5 indicate liked items, while ratings below 5 indicate disliked items.

\subsubsection{Evaluation Metrics}
In line with recent studies, we adopt two commonly used evaluation metrics: Area Under the ROC Curve (AUC) and Binary Cross-Entropy (BCE) loss \cite{liu2020general, chen2021autodebias, liu2023bounding}. The AUC metric assesses the discriminative power of an algorithm in distinguishing between positive and negative classes. It can be interpreted as the probability that the algorithm predicts a random positive sample with a higher score than a random negative sample. A higher AUC value indicates better performance in ranking positive samples higher than negative samples. On the other hand, the BCE loss considers the model's confidence in its predictions. It penalizes the model heavily when it assigns a wrong label to a sample with high confidence. The model is encouraged to make confident and accurate predictions by minimizing the BCE loss.
\begin{table}[t]
\caption{Statistics of Datasets}
\label{table:dataset_statistics}
\begin{center}
\begin{tabular}{l|cccc}
\toprule
Dataset & $|D^r|$ & $|D^b|$ & $\frac{|D^r|}{|D^b|}$ & PR$(D^r, D^b)$ \\ \midrule  
Coat & 4640 & 6594 & 0.70 & (0.05, 0.09)\\ 
YahooR3 & 54000  & 311704 & 0.17 & (0.03, 0.24)\\
 \bottomrule
\end{tabular}
\end{center}
\end{table}

\subsubsection{Baselines}
Recent studies predominantly rely on matrix factorization, while our framework leverages neural networks due to their enhanced representational capacity, which has made them the primary architecture in state-of-the-art recommendation algorithms \cite{cheng2016wide,he2017neural,xue2019deep}. To ensure a fair comparison, we consider existing studies that can be adapted as loss functions similar to our proposed EnG loss. These include the Bridge and Refine strategies \cite{liu2020general}, Autodebias approach \cite{chen2021autodebias}, and the DUB method \cite{liu2023bounding}. Two baseline models, "Uniform" and "Union," are also included for performance comparison. The "Uniform" model is exclusively trained on uniformly collected data $D^r$, while the "Union" model is trained on the entire dataset $D$, including both uniformly collected and biased logged data.

Hyperparameters were selected based on the achieved AUC scores on the validation set. The range for the regularization weights, including $\gamma_{\text{reg}}$, was set as [1e-4, 1e-3, 1e-2, 1e-1, 1]. The batch size of the data loader was varied among the values [16, 32, 64, 128, 256]. The hyperparameter $\alpha_{\text{refine}}$ was explored within the range [1e-4, 1e-3, 1e-2, 1e-1, 0.1, 0.2, ..., 0.9]. For the YahooR3 dataset, embedding dimensions of [10, 20, 50, 100, 200] were used for user and item IDs. Network architectures with both 2-layer and 3-layer configurations were examined, with neuron counts per layer ranging from [32, 64, 128, 256, 512], ensuring that the teacher network had significantly fewer parameters than the student network. Results on the Coat dataset were averaged over 10 runs, while results on the YahooR3 dataset were averaged over 5 runs. Convergence was ensured using an early stopping method based on the BCE score on the validation set. The code will be made available upon acceptance of the paper.

\subsubsection{Performance Investigation using conventional training schema}
In this section, similar to prior studies, we employ conventional training schema to evaluate the efficacy of the EnG framework in utilizing limited uniformly collected data. The uniform portion of the dataset was divided into three disjoint sets: training, validation, and testing. The validation and test sets were chosen explicitly from the uniformly collected data to ensure an unbiased evaluation process. For both datasets, 20 percent of the uniformly collected data was allocated for training, while the remaining uniform data were evenly distributed between the validation and test sets. The experimental results are presented in Table \ref{table:classic_result}.

\begin{table}[t]
\caption{Performance Comparison - Conventional Training}
\label{table:classic_result}
\begin{center}
\begin{tabular}{l|cc|cc}
\toprule
\multirow{2}{*}{Method} & \multicolumn{2}{c|}{Coat}       & \multicolumn{2}{c}{YahooR3}    \\  
                        & AUC            & BCE            & AUC            & BCE            \\ \midrule
EnG-MAE                 & 0.829 & 0.174          & 0.775          & 0.132          \\
EnG-MSE                 & 0.829 & 0.171 & 0.775          & 0.192          \\
EnG-KL                  & 0.835 & 0.164          & 0.780          & 0.228          \\
EnG-Jeffrey                  & \textbf{0.837} & \textbf{0.160}          & \textbf{0.787} & 0.209          \\ \midrule
Bridge                  & 0.801          & 0.183          & 0.668          & 0.224 \\
Refine                  & 0.774          & 0.209          & 0.765          & 0.325          \\
Autodebias              & 0.779   & 0.242          & 0.748          & \textbf{0.103} \\
DUB                     & 0.796          & 0.194          & 0.727          & 0.112          \\ \midrule
Union & 0.749 & 0.204 & 0.631 & 0.279 \\
Uniform                 & 0.676          & 0.186          & 0.614 & 0.112 \\ \bottomrule
\end{tabular}
\end{center}
\end{table}
Upon analyzing the results, several noteworthy observations emerge. First, strategies that utilize uniformly collected data effectively yield higher AUC scores than the Union strategy, which combines all available data. Although all strategies successfully use this information, the proposed EnG framework exhibits significantly higher AUC scores than other state-of-the-art methods. This indicates its superior effectiveness in leveraging the information from uniformly collected data. It is worth noting that the YahooR3 dataset, being sparser and more challenging, presents more considerable disparities in the positive ratio feedback distribution (PR) between uniform and biased data when compared to the Coat dataset (refer to Table \ref{table:dataset_statistics}). This more significant divergence poses a heightened challenge in effectively utilizing the uniformly collected data within the YahooR3 dataset. Consequently, the AUC scores achieved for the Coat dataset consistently surpass those obtained for the YahooR3 dataset.
Furthermore, when examining specific strategies, it is observed that the AUC scores of the Bridge and DUB methods on the YahooR3 dataset are notably lower than their respective AUC scores on the Coat dataset. The Bridge strategy, where teacher and student networks are trained simultaneously, encounters difficulty achieving stability during training. This, combined with the challenge of extracting knowledge from the uniformly collected data in the YahooR3 dataset, explains the performance drop of the Bridge strategy. In the case of the DUB loss function, the term that aims to make the student mimic the teacher's prediction error on uniform data becomes less effective due to the more significant discrepancy between the feedback distributions of uniform and biased data in the YahooR3 dataset.
Lastly, the BCE score is directly influenced by the similarity between the training and test distributions, as it measures the discrepancy between predicted and ground truth distributions. Consequently, the Autodebias approach achieves a lower BCE score as it directly optimizes its parameters using uniformly collected data without employing a teacher-student architecture. However, since the difference in the positive ratio distribution between uniform and biased data is much more significant in the YahooR3 dataset compared to Coat, incorporating biased data can help reduce the BCE score on the Coat dataset. As a result, the proposed EnG framework achieves a lower BCE score on the Coat dataset.

\subsubsection{Performance Investigation using sequential training schema}
In this section, we adopt a sequential training schema described in Section \ref{section:sequential_trainnig_schema}. The training data is divided into 20 batches, and we carefully adjust the selection ratio ($\rho$) and the proportion of uniformly collected data used in training to ensure that the ratio of unbiased data to biased data in each batch remains low (approximately 5 percent).

The experimental results are presented in Table \ref{table:sequential_training}. We observe a decrease in AUC scores and an increase in BCE scores compared to the results of previous experiments. This can be attributed to two main reasons. First, the reduced amount of uniformly collected data poses challenges for the methods to extract unbiased information, leading to a decline in performance. Second, adopting a sequential training schema intensifies the self-feedback loop effect, where the model's recommendations are increasingly influenced by its previous predictions, potentially introducing bias in subsequent training iterations.
Consistent with the previous results, except for the DUB strategy on the Coat dataset, the EnG framework and state-of-the-art methods achieve higher AUC scores than the Union strategy. However, the EnG framework demonstrates further improvements in AUC scores, indicating its capability to effectively utilize smaller amounts of uniformly collected data, even in a sequential training schema. This highlights the effectiveness of the EnG methods in leveraging limited amounts of unbiased data for improved recommendation performance. The observed decrease in AUC score for the DUB strategy on the Coat dataset, resulting in inferior performance compared to the Union strategy, can be attributed to the specific characteristics of the DUB approach. The DUB strategy incorporates a loss term that encourages the student network to emulate the prediction errors of the teacher network on uniformly collected data. However, in this scenario where the teacher network's reliability is diminished, this loss term may introduce misleading guidance to the DUB model, thereby hindering its overall performance. Lastly, the intensified bias is evident through higher BCE scores for all methods, including Autodebias, on the YahooR3 dataset. The Uniform strategy attains the lowest BCE score, benefiting the most from the similarity between the training and test data distributions. However, relatively smaller BCE scores are observed for the EnG methods on the Coat dataset. The reasons for this can be attributed to similar explanations as before.
\begin{table}[t]
\caption{Performance Comparison - Sequential Training}
\label{table:sequential_training}
\begin{center}
\begin{tabular}{l|cc|cc}
\toprule
\multirow{2}{*}{Method} & \multicolumn{2}{c|}{Coat}       & \multicolumn{2}{c}{YahooR3}    \\  
                        & AUC            & BCE            & AUC            & BCE            \\ \midrule
EnG-MAE     & 0.790          & 0.220          & 0.739          & 0.132 \\
EnG-MSE     & 0.787          & 0.219          & 0.758          & 0.202 \\
EnG-KL      & \textbf{0.797} & \textbf{0.191} & 0.760          & 0.255 \\
EnG-Jeffrey      & 0.795          & 0.216          & \textbf{0.769} & 0.231 \\ \midrule
Bridge      & 0.737          & 0.280          & 0.653          & 0.226 \\
Refine      & 0.708          & 0.296          & 0.731          & 0.359 \\
Autodebias  & 0.740          & 0.323          & 0.691          & 0.333 \\
DUB         & 0.644          & 0.207 & 0.646          & 0.128 \\ \midrule
Union & 0.687 & 0.209 & 0.612 & 0.289 \\
Uniform     & 0.594          & 0.281          & 0.547          & \textbf{0.114} \\ \bottomrule
\end{tabular}
\end{center}
\end{table}

\subsubsection{Impact of Introducing Thompson Sampling}
In this section, we aim to investigate the potential performance enhancement of our proposed model through the incorporation of Thompson sampling (TS) and the introduction of exploratory behavior in the recommendation process. We analyze two settings, differing only in applying Thompson sampling while selecting the model's training data. The findings of this analysis are summarized in Table\ref{table:thompson-sampling}. 
Initially, we observe that the utilization of Thompson sampling improves both the AUC and BCE scores of the EnG methods. It is important to note that the architectural design and hyperparameters remain consistent across both methods. Thus, the observed improvement can be directly attributed to the enhanced quality of the training data resulting from the application of Thompson sampling. Furthermore, our analysis reveals that the improvements in terms of the BCE score are more pronounced compared to the enhancements in the AUC score.  This could be attributed to utilizing dropout techniques during inference in the Thompson Sampling models. This leads to recommending items with lower scores if dropout techniques were not employed. This, in turn, increases the inclusion of uncertain items in the model's training data for the subsequent training iteration. 
Incorporating these uncertain items contributes to enhancing both coverage and diversity within the training data. Given that the BCE score is sensitive to the dissimilarity between the training and test data distributions, including these uncertain items helps mitigate this discrepancy, thereby reducing the BCE score. 
Lastly, the improvement in AUC for the MAE and Jeffreys regularization loss functions is more substantial compared to the MSE and KL regularization loss functions. This observation can be attributed to the nature of the loss functions themselves. The MAE loss imposes stricter constraints than MSE loss on the output range of the teacher and student networks, as it operates within the range of 0 to 1. Similarly, the Jeffreys loss, a symmetric KL variant, is also a more restrictive loss function. Consequently, adding exploration has a more significant potential to enhance the performance of models trained with MAE and Jeffreys regularization loss functions.

\begin{table}[t]
\caption{Impact of Thompson Sampling}
\label{table:thompson-sampling}
\begin{center}
\begin{tabular}{l|cc|cc}
\toprule
\multirow{2}{*}{Method} & \multicolumn{2}{c|}{Coat}       & \multicolumn{2}{c}{YahooR3}    \\  
                        & AUC            & BCE            & AUC            & BCE            \\ \midrule
EnG-MAE         & 0.726 & 0.266 & 0.665 & 0.175    \\
EnG-MAE(TS)     & \textbf{0.777} & \textbf{0.192} & \textbf{0.728}& \textbf{0.105}    \\ \midrule
EnG-MSE         & 0.740 & 0.260 & 0.734 & 0.198 \\
EnG-MSE(TS)     & \textbf{0.779} & \textbf{0.180} & \textbf{0.741}& \textbf{0.116}   \\ \midrule
EnG-KL          & 0.740 & 0.268 & 0.763 & 0.216    \\
EnG-KL(TS)      & \textbf{0.783} & \textbf{0.110} & \textbf{0.768}& \textbf{0.110}   \\ \midrule
EnG-Jeffrey         & 0.715 & 0.263 & 0.672 & 0.201   \\
EnG-Jeffrey(TS)     & \textbf{0.773} & \textbf{0.190} & \textbf{0.773}& \textbf{0.111} \\ \bottomrule
\end{tabular}
\end{center}
\end{table}

\subsubsection{Impact of Regularization Losses}
The findings presented in Table \ref{table:classic_result}, Table \ref{table:sequential_training}, and Table \ref{table:thompson-sampling} demonstrate that the KL and Jeffreys loss functions exhibit a slight advantage over the MAE and MSE loss functions. This performance discrepancy can be attributed to the inherent properties of the KL and Jeffreys distances, which incorporate a logarithmic penalty to measure the discrepancy between the teacher and student network predictions. As a result, these loss functions impose a more substantial penalty on the divergence between the output probability distributions of the two networks. Consequently, they facilitate a more effective transfer of knowledge from the teacher to the student network, ultimately enhancing the performance of the student model.

\subsubsection{Ablation Study}
In this section, we investigate the impact of various learning parameters on the performance of our proposed EnG framework. Specifically, we analyze the effects of different factors, namely the amount of uniformly collected data used in training ($|D^r_{\text{train}}| / |D^r|$), the selection ratio ($\rho$) for choosing top predictions from each training batch, and the dropout ratio that determines the level of exploration during prediction.

\begin{figure}[t]
\begin{center}
\begin{tabular}{cc}
  \includegraphics[width=0.5\linewidth]{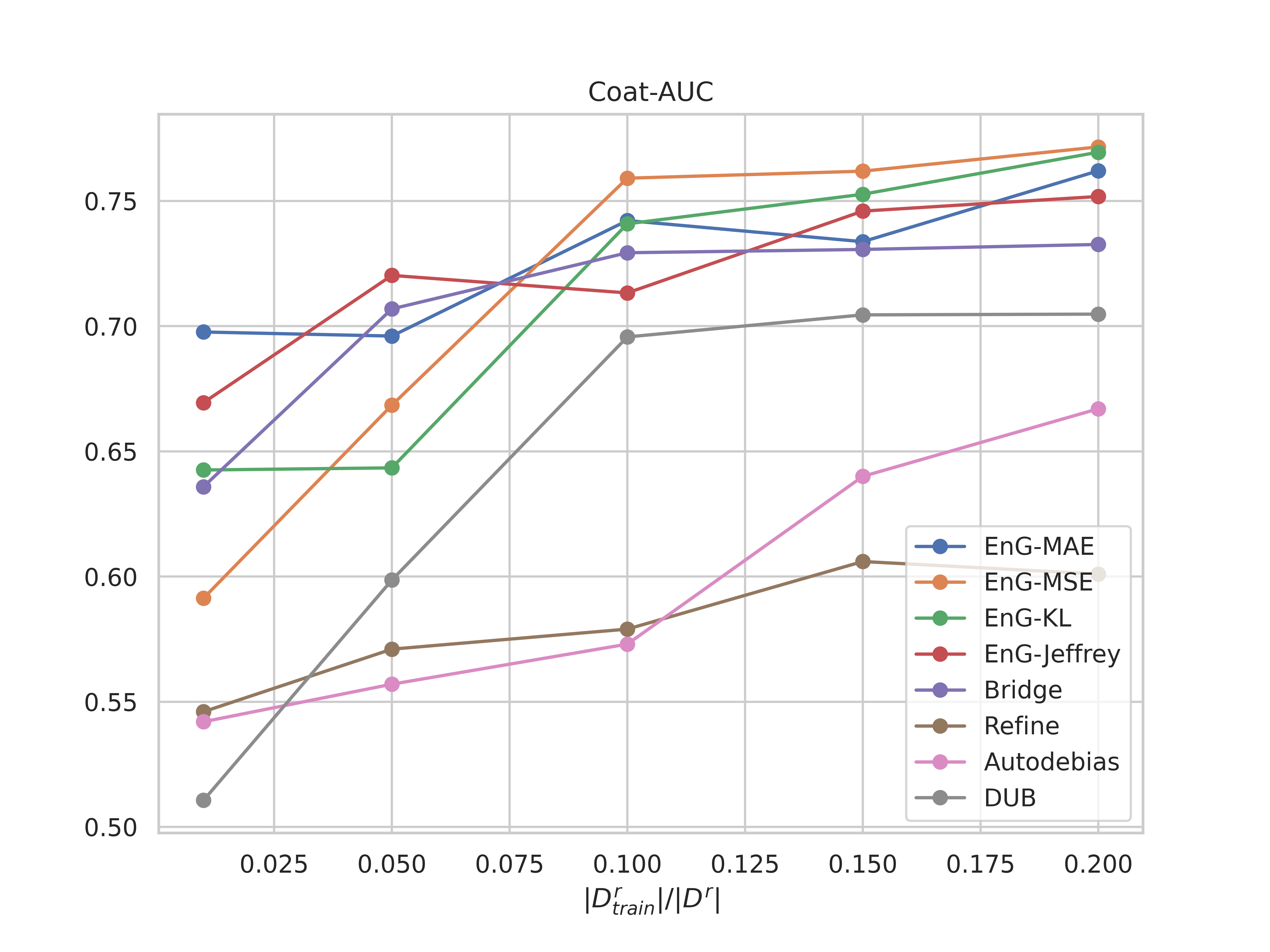} &   \includegraphics[width=0.5\linewidth]{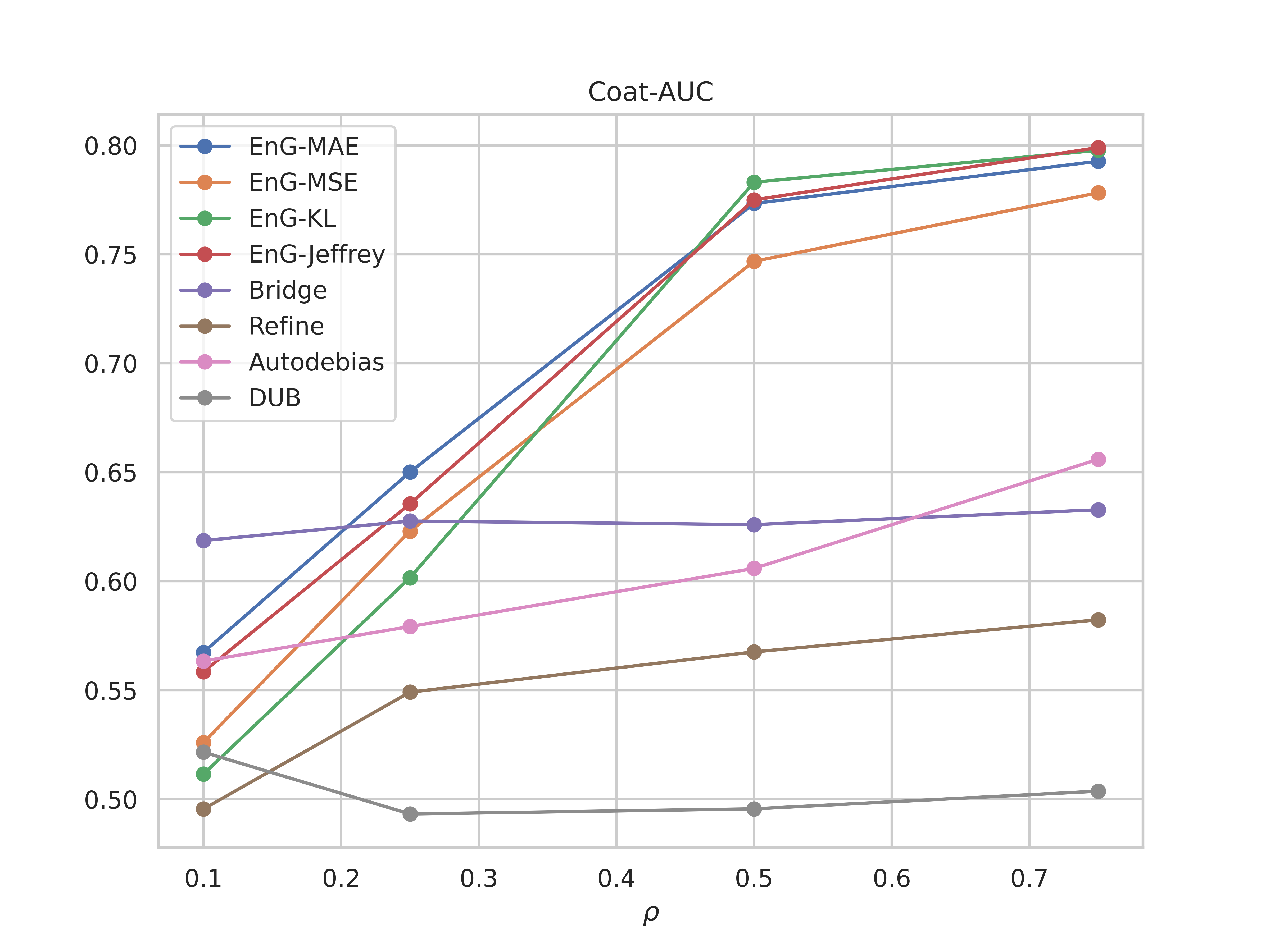} \\
 \includegraphics[width=0.5\linewidth]{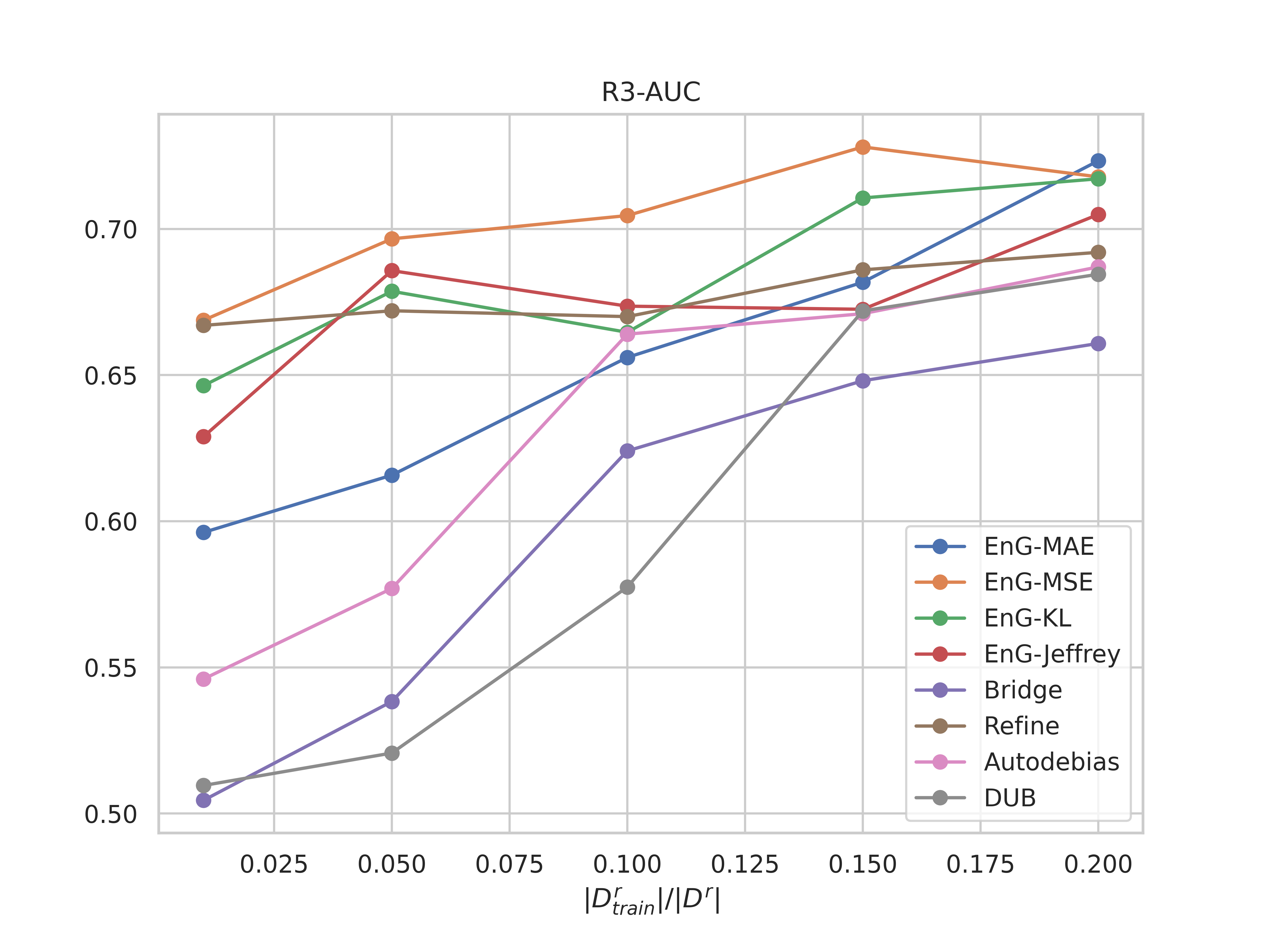} &   \includegraphics[width=0.5\linewidth]{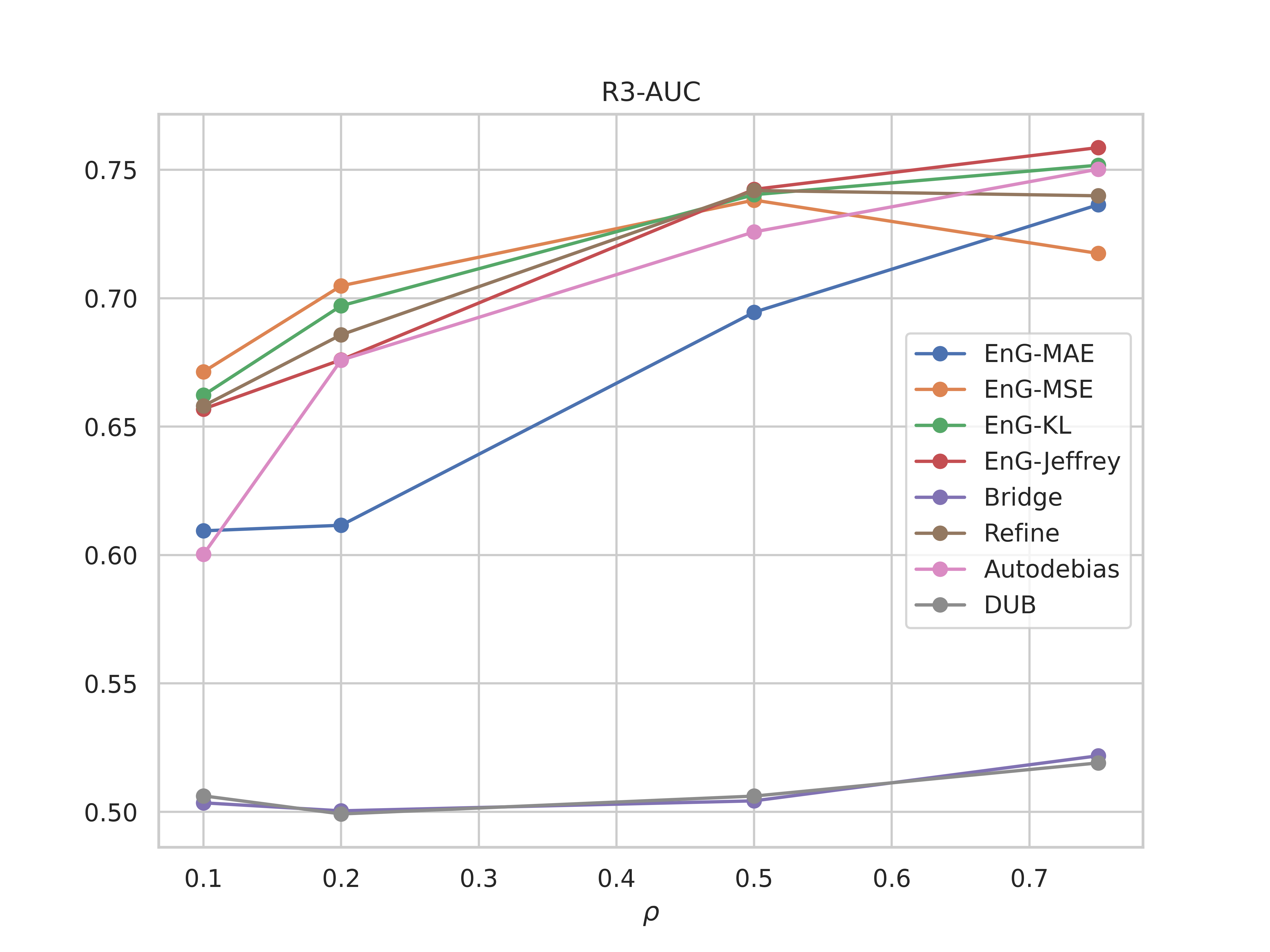} \\
\end{tabular}
\end{center}
\caption{Varying different sequential training parameters.}
\label{fig:ablation-unifrom-selection}
\end{figure}

The influence of using different amounts of uniformly collected data and varying selection ratios ($\rho$) is presented in Fig \ref{fig:ablation-unifrom-selection}. To solely consider EnG's capability to extract information from uniformly collected data, we restrict the amount of biased data by setting $\rho=0.25$ for the Coat dataset and $\rho=0.1$ for the YahooR3 dataset. Our observations reveal that the AUC scores generally increase as more uniformly collected data is available in the training dataset. Furthermore, the EnG methods exhibit promising performance in utilizing uniformly collected data compared to state-of-the-art competitive methods. Notably, the Refine strategy demonstrates limited effectiveness in leveraging uniformly collected data, potentially due to the manner in which uniform data is incorporated. In Refine strategy, the teacher network imputes labels on biased data, but given the small amount of uniformly collected data, these imputed labels could not effectively transfer the knowledge from the uniform data.

In order to exclusively examine the impact of selecting varying amounts of biased data in each training batch, we limit the quantity of uniform training data to 1 percent of the total available uniform data within the datasets. The results demonstrate that all methods exhibit an increasing trend in AUC scores with a larger quantity of biased data selected in each training batch (Fig \ref{fig:ablation-unifrom-selection} right). Notably, the EnG methods show further improvements with an increased amount of biased data. However, a significant performance gap is observed between the EnG methods and other state-of-the-art approaches on the Coat dataset. This can be attributed to two reasons: first, the EnG framework effectively utilizes small quantities of uniformly collected data, as observed in previous experiments, and second, the relatively smaller disparity in positive ratios between the uniform and biased data in the Coat dataset, as compared to the YahooR3 dataset, provides an explanation for the effectiveness of EnG methods in utilizing a larger quantity of biased data in conjunction with limited unbiased data. Notably, this increasing trend is not observed in the Bridge and DUB methods. In the Bridge strategy, where the teacher and student networks are trained simultaneously, the interleaving training procedure becomes detrimental when only a small portion of uniform data is used. Similarly, enforcing the student network to mimic the teacher's prediction error in the DUB strategy can lead to adverse effects when the teacher network is relatively weak.

\begin{figure}[t]
\begin{center}
\begin{tabular}{cc}
  \includegraphics[width=0.5\linewidth]{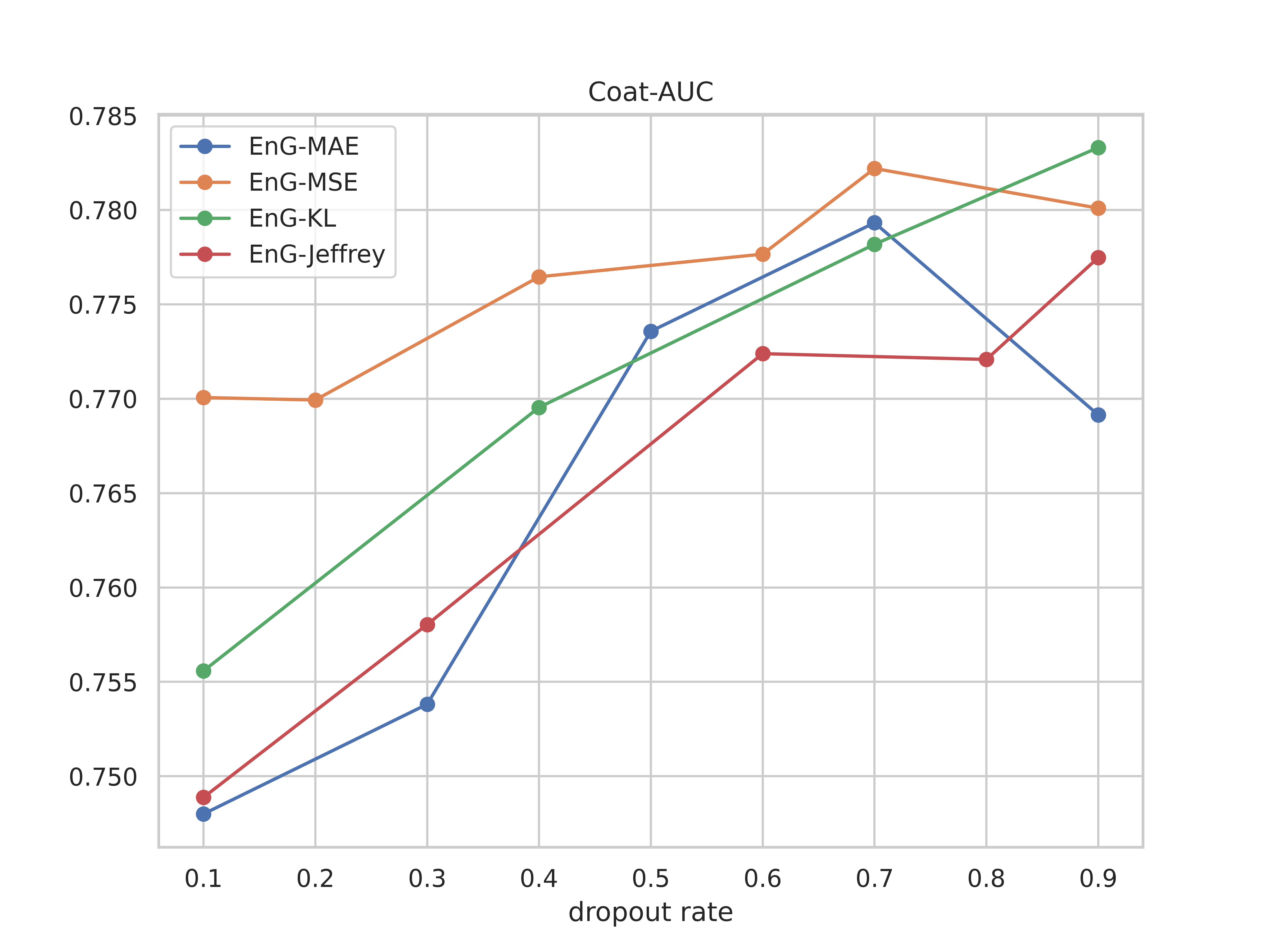} &   \includegraphics[width=0.5\linewidth]{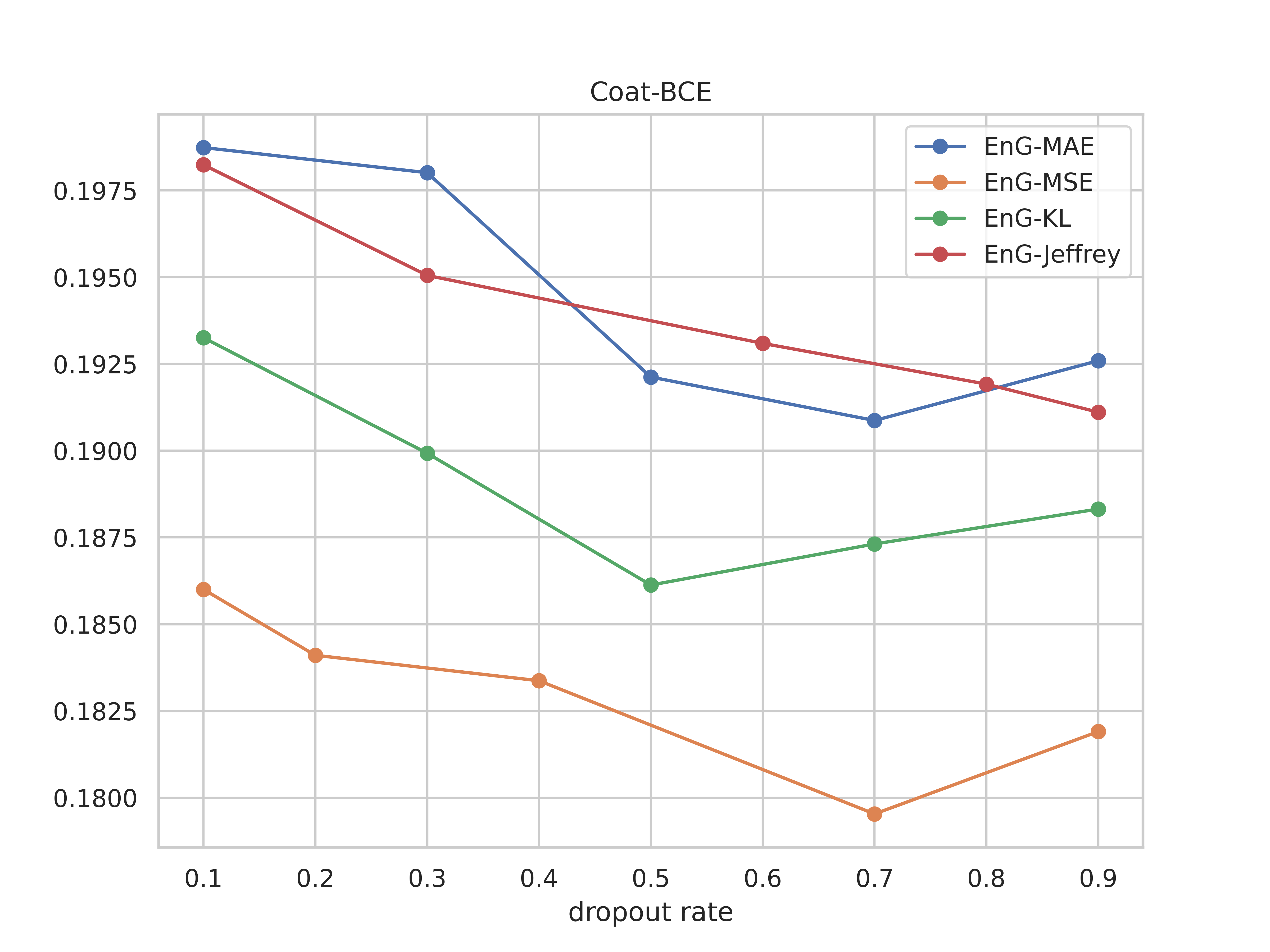} \\
 \includegraphics[width=0.5\linewidth]{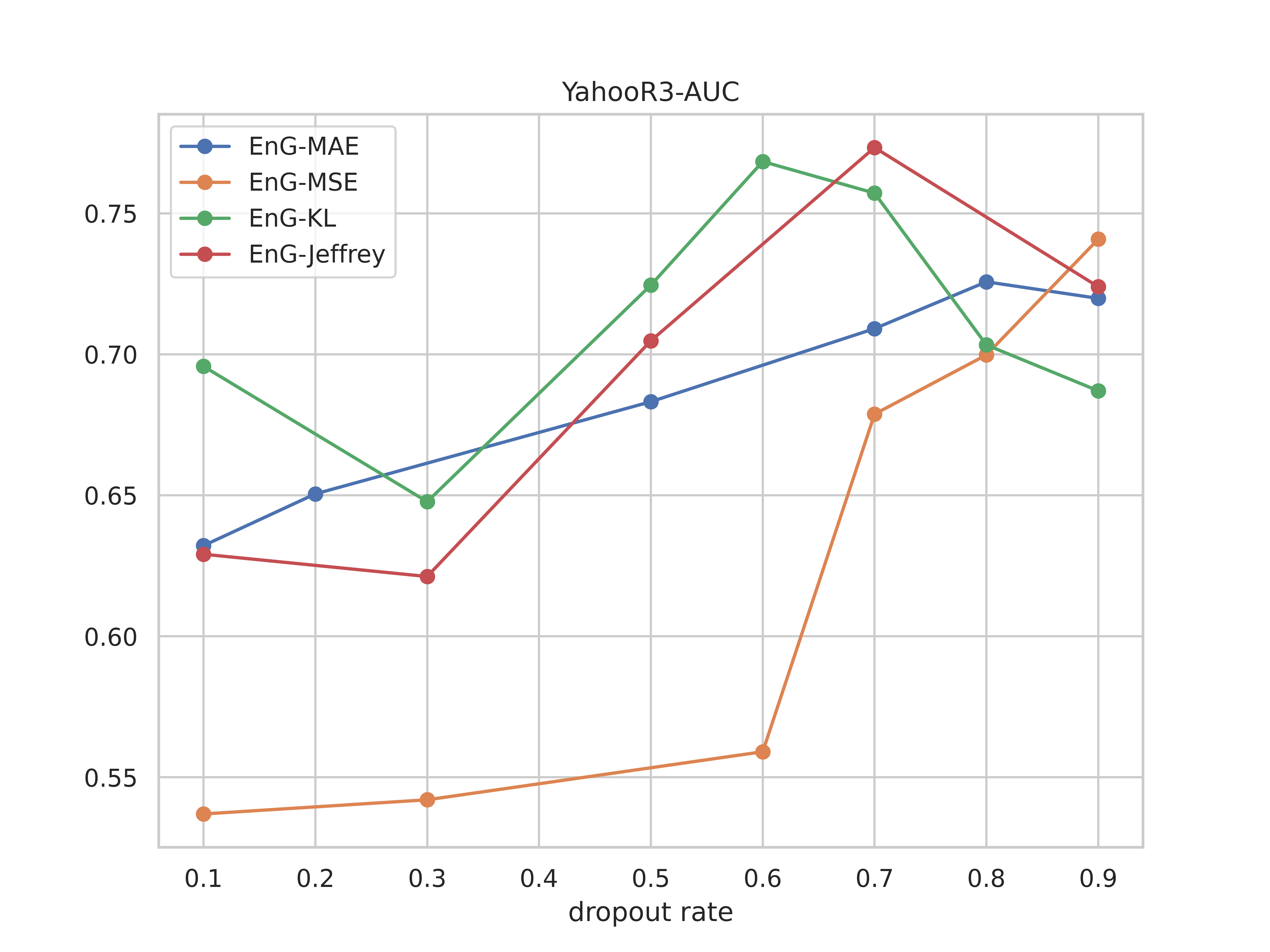} &   \includegraphics[width=0.5\linewidth]{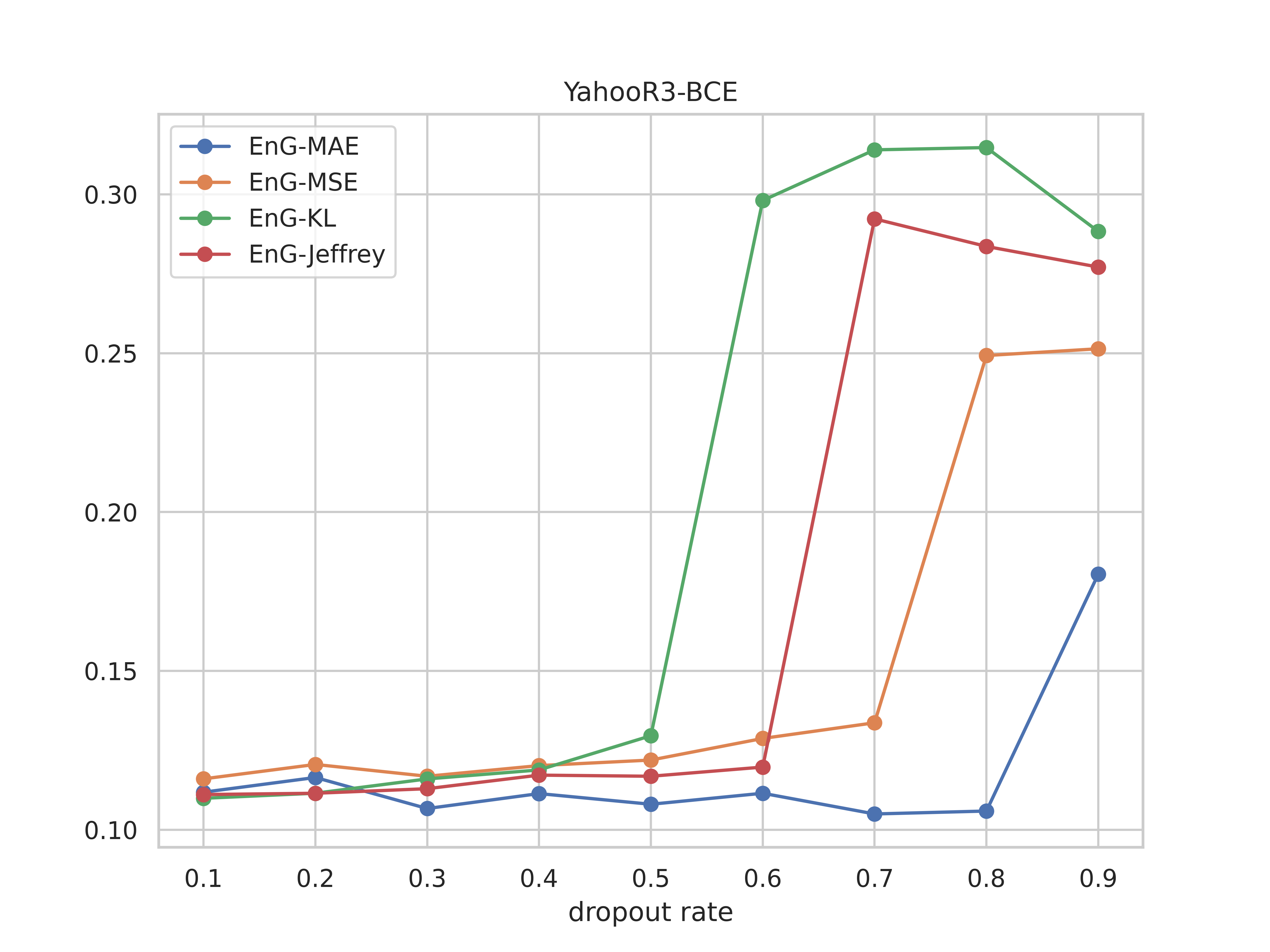} \\
\end{tabular}
\end{center}
\caption{Impact of varying dropout rate on EnG framework.}
\label{fig:ablation-dropout}
\end{figure}

The increase in dropout rate introduces additional randomness in the outputs of neural networks, leading to increased exploration and stronger regularization during training. However, excessively high dropout rates can have detrimental effects on model performance due to excessive exploration and regularization, hindering the model's ability to exploit learned knowledge effectively. Therefore, it is crucial to strike a balance when incorporating randomness into the model. In Fig \ref{fig:ablation-dropout}, we analyze the impact of varying dropout rates on the performance of EnG methods. The AUC scores demonstrate an upward trend as the dropout rate increases, indicating that a moderate increase in the dropout rate can benefit the methods by encouraging exploration and appropriate regularization. However, there is a trade-off between exploration and exploitation, as some methods initially benefit from increasing dropout rates but experience a decline in performance when the dropout rate becomes excessively high. Similar trends are observed in the BCE scores. For example, on the Coat dataset, the BCE scores initially decrease with an increasing dropout rate but eventually rise due to excessive exploration and regularization. Conversely, on the YahooR3 dataset, lower dropout rates yield better performance in BCE scores, with an increase in the dropout rate leading to higher BCE scores.
\section{Conclusion and Future Works}
In conclusion, recommendation systems commonly suffer from bias induced by the self-feedback loop, which arises from continuous training of the algorithm on its previous recommendations. To address this issue, we propose a teacher-student architecture that effectively leverages a small quantity of uniformly collected data to learn an unbiased model. By incorporating Thompson sampling, we ensure that the architecture learns an unbiased recommendation system and generates recommendations that contribute to improved training data for subsequent training iterations. This integration allows the model to exhibit exploratory behavior towards items it is unaware of, resulting in less biased training data for future training iterations. To evaluate the effectiveness of our architecture, we introduce a sequential training schema that emulates the continuous training process observed in real-world recommendation systems.

Distribution shift poses a significant challenge in recommendation systems, as the distributions of user interests, item preferences, and contextual information undergo dynamic changes over time. When combined with the continuous training nature of recommendation systems, this distribution shift can exacerbate the self-feedback bias. Despite the importance of this dynamic nature in recommendation problems, current studies and datasets do not adequately capture it. Hence, investigating the relationship between distribution shift and the self-feedback loop represents a promising avenue for future research.

\section*{References}
\printbibliography[heading=none]
\end{document}